# Analyzing covert social network foundation behind terrorism disaster

## Yoshiharu Maeno

Graduate School of Systems Management, Tsukuba University, Otsuka 3-29-1, Bunkyo-ku, Tokyo 112-0012, Japan

E-mail: maeno.yoshiharu@nifty.com

## Yukio Ohsawa

School of Engineering, University of Tokyo, Hongo 7-3-1, Bunkyo-ku, Tokyo 113-8563, Japan

E-mail: ohsawa@q.t.u-tokyo.ac.jp

Abstract: This paper addresses a method to analyze the covert social network foundation hidden behind the terrorism disaster. It is to solve a node discovery problem, which means to discover a node, which functions relevantly in a social network, but escaped from monitoring on the presence and mutual relationship of nodes. The method aims at integrating the expert investigator's prior understanding, insight on the terrorists' social network nature derived from the complex graph theory, and computational data processing. The social network responsible for the 9/11 attack in 2001 is used to execute simulation experiment to evaluate the performance of the method.

*Y. Maeno and Y. Ohsawa*

## 1 Introduction

Terrorism is a man-made disaster. It causes great economic, social and environmental impacts. It is different from the emergence arising from natural disasters (earthquakes, hurricanes etc.), in that active non-routine responses are always necessary as well as the disaster recovery management. The short-term target of the responses includes interpretation of the hidden intention of the terrorism and arrest of the terrorists responsible for the disaster. The long-term target is identification and weakening of the covert foundation which raises, encourages, and helps terrorists. For example, a conspirator, named Mustafa A. Al-Hisawi, had attempted to help terrorists enter the United States (according to Wikipedia), and provided Mohamed Atta and the hijackers responsible for the 9/11 attack in 2001 with financial support worth more than $300,000 (according to New York Times). Future terrorism disasters are mitigated and eliminated by dismantling such a covert social network foundation existing behind the terrorism.

This paper addresses a method to analyze the covert social network foundation existing behind the terrorism disaster. Mathematically, the objective of the analysis is to solve a node discovery problem. The problem means to discover a node, which functions relevantly in a complex social network, but escaped from monitoring on the presence and mutual relationship of nodes either intentionally or accidentally. Practically, the problem is difficult to solve because of the 2 reasons. First, the terrorism disaster is infrequent and non-routine, and consequently does not take a fixed form. Second, the intelligence and surveillance (communication logs and meeting records are examples) on the covert social network is limited, or still worse, missing completely. We can not, therefore, rely on conventional machine learning and probabilistic inference techniques under such a condition. Instead, our method aims at integrating the expert investigator's prior understanding, insight on the terrorists' social network nature derived from the complex graph theory, and computational data processing.

The approach to solve the node discovery problem is developed in section 2. The social network of the hijackers and conspirators in the 9/11 attack is reviewed in section 3. In section 4, the network in section 3 is used to execute simulation experiment to discover a covert conspirator by the approach presented in section 2. Related works are summarized in section 5. Concluding remarks are presented in section 6.

## 2 Approach

### 2.1 Problem definition

Before presenting our approach, we define the node discovery problem and describe assumptions. The node discovery problem in a complex network is new in two senses.

*Analyzing covert social network foundation behind terrorism disaster*

First, the problem has not attracted much attention from researchers. It is in contrast to that a link discovery problem is studied to predict unknown chemical reaction between 2 molecules in bio-informatics intensively. Second, the nature of covert social network foundation behind the terrorism is not understood well, despite the fact that many organizations and human relationships are described by scale-free networks or small worlds.

This problem is illustrated in Figure 1. The inset (a) represents the observed records on the organization under investigation. Geographically distributed persons are likely to use the Internet to join the organizational decision-making process, to determine the attack plan, and to give instructions to the terrorists. In the example, the records are sets of participants of email-based on-line group discussions. Four persons ($p_0$, $p_1$, $p_2$, $p_3$) joined the first discussion (subject 0). We can gather a number of records automatically if we assume simply that an individual discussion is indicated by the same email subject. The records are in the form of a market basket shown by eq.(1).

$$b_i = \{p_j\} \quad (0 \leq i \leq |b|-1) \quad . (1)$$

The order of records and the order of persons in a record are not significant here. The problem may be extended to a time-sensitive or causality-sensitive situation where the orders provide us with a significant clue to solve the problem. Such a situation is for future study. Cluster structures can be extracted from the records. The cluster is a group of persons, between whom communication is active. In the example, two clusters $c_0$ ($p_0$, $p_1$, $p_2$, $p_3$), and $c_1$ ($p_4$, $p_5$, $p_6$, $p_7$) can be extracted. They are visualized on the social network diagram in the blue box. The diagram is an undirected graph. The black nodes denote persons, and black links between the nodes denotes the presence of active communication. The links are drawn according to the degree of activeness between two nodes at the end of the link. The links are not directed because the communication is bi-directional. The cluster is not necessarily a clique (a complete graph where links exist between every possible pair of the nodes).

The inset (b) represents the latent structure behind the observed records. In this example, the latent structure is a covert participant (or participants) who used telephone to tell persons in the separate clusters to encourage the organization-wide communication, and to adjust the direction of decision. The person escaped from the email surveillance in (a). The fifth record in (a) is not consistent from the viewpoint of the overall cluster structure of the organization. This is a clue. The unobserved person $p_x$ may be hidden in the empty space between the gateway persons ($p_0$, $p_4$) in the clusters. The person is indicated by a red node and red links connecting the clusters in the social network diagram. The red node is a hypothetical candidate of the latent structure.

Our aim is to reveal clues to infer (b) from (a). Note that the identity of the red nodes can not be derived from the observed records in (a) automatically, but is inferred with the aid of the expert investigator's knowledge. Our primary interest here lies in drawing a social network diagram to invent hypothesis on the latent structure which is ready for testing. The interactive process for this purpose is presented in the following.

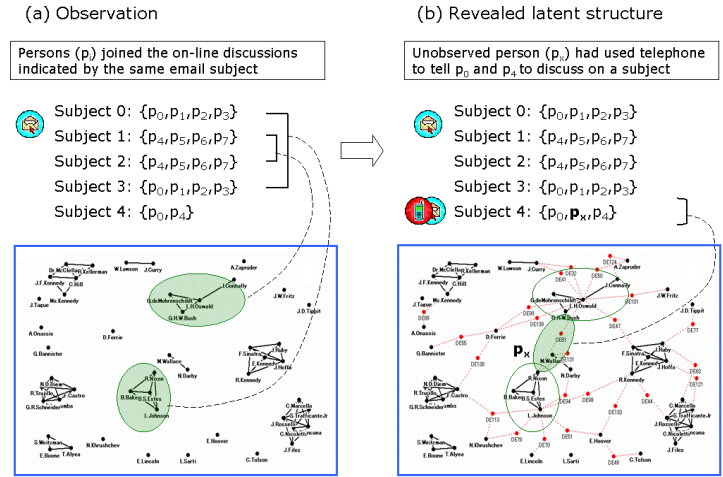

Figure 1 Inset (a) represents the observed records on the participants ($p_i$) of email-based on-line group discussions. A record is a list of persons who join an individual discussion indicated by the same email subject. Two clusters $c_0$ ($p_0$, $p_1$, $p_2$, $p_3$), and $c_1$ ($p_4$, $p_5$, $p_6$, $p_7$) can be extracted from the five records. Black nodes denote persons. Black links between the nodes denotes the presence of communication. Inset (b) represents the latent structure (a covert participant) behind the observed records. The fifth record is a clue to infer that an unobserved person $p_x$ may be hidden in the empty space between the gateway persons ($p_0$, $p_4$) in the clusters. The unobserved person may use telephone to foster communication between the gateway persons. Our aim is to reveal (b) from (a).

## 2.2 Interactive process

We propose an interactive process starting from the intelligence, surveillance, and the prior knowledge of expert investigators toward the hypothesis on the latent structure. Figure 2 shows the process. The algorithm, used in the computational data processing shown in the dashed grey box, visualizes the observed records on communication in the form of eq.(1) into a social network diagram. It consists of clustering and ranking procedure. The clustering procedure evaluates the activeness of communication between the persons, and uses the prior knowledge such as the number of groups or the known group leaders. The ranking procedure calculates likeliness of the suspicious inter-cluster relationships, which originates in the unobserved person hidden in the empty spots between the clusters, and indicates the position of the person as a red node.

The expert investigators explore the difference between the visualized social network diagram and the prior understanding. The difference is expected to be a trigger to notice something new. The expert can update the prior understanding, iterate the above procedures, and finally invent a hypothesis on the latent structure (Maeno, 2007). The details of the algorithm are presented in the following. The essence of the algorithm is the ranking function to calculate likeliness of the suspicious inter-cluster relationships.

*Analyzing covert social network foundation behind terrorism disaster*

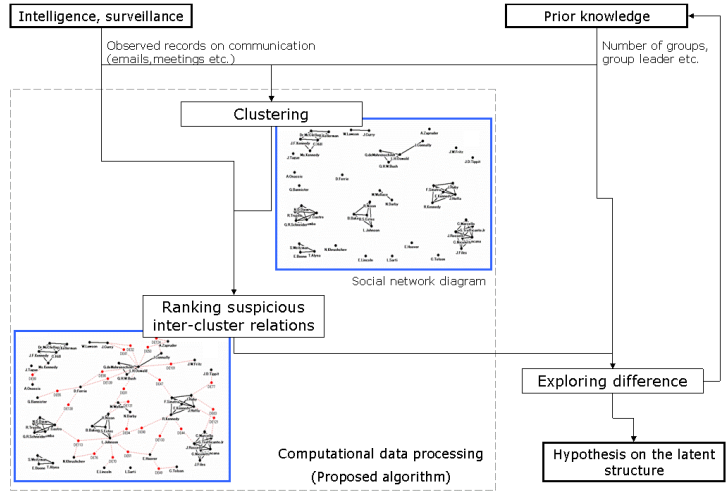

Figure 2 Interactive process from the intelligence, surveillance and prior knowledge of the expert investigators toward the hypothesis on the latent structure. The computational data processing in the dashed grey box visualizes the observed records on communication in the form of eq.(1). It consists of clustering using the prior knowledge, and ranking of suspicious inter-cluster relationships which originates in the unobserved person. The expert explores the difference between the visualized social network diagram and the prior understanding, which is the basis to invent a hypothesis.

## *2.3 Computational data processing*

Our algorithm focuses on inter-cluster relationships in a social network (Ohsawa, 2005). Examples of the inter-cluster relationships include sharing of information on the guard system among the hijacker groups via a conspirator, or efficient multicast of a directive to the groups from a conspirator. The input of the algorithm is the observed records in eq.(1). The output is the ranking of the individual records (indicating suspicious inter-cluster relationships or unobserved persons playing a catalyst role among the clusters), and the persons in the clusters playing a gateway role to the unobserved person. The output is further processed to draw a social network diagram.

As a preparation, we define a simple Boolean function $B(s)$ by eq.(2). It returns 1 if the statement s is true, and 0 otherwise.

$$B(s) = \begin{cases} 1 & \text{if } s \text{ is true} \\ 0 & \text{otherwise} \end{cases} \quad (2)$$

At first, the all persons appearing in the observed records $b_i$ in eq.(1) are grouped into clusters $c_j$. The number of clusters $|c|$ depends on the prior knowledge. Mutually close persons form a cluster. The measure of closeness between a pair of persons is evaluated by Jaccard's coefficient. It is defined by eq.(3). The function $F(p_i)$ is the occurrence frequency of a person $p_i$ in the records. The closeness means activeness of the communication if the record is a set of the persons appearing together in the emails, conversations, or meetings. Jaccard's coefficient is used widely in link discovery, web mining, or text processing.

$$J(p_i, p_j) = \frac{F(p_i \cap p_j)}{F(p_i \cup p_j)} = \frac{\sum_{0 \leq k \leq |b|-1} B((p_i \in b_k) \wedge (p_j \in b_k))}{\sum_{0 \leq k \leq |b|-1} B((p_i \in b_k) \vee (p_j \in b_k))}. \quad (3)$$

Here, we employ the k-medoids clustering algorithm (Hastie, 2001). It is an EM (expectation-maximization) algorithm similar to the k-means algorithm for numerical data. A medoid $p_{\text{med}}(c_j)$ locates most centrally within a cluster $c_j$. It corresponds to the center of gravity in the k-means algorithm. The modoid persons are selected at random initially. The other |p|-|c| persons are classified into the clusters whose medoids is the closest. A new medoid is selected within an individual cluster so that the sum of Jaccard's coefficients between the modoid and persons in the cluster can be maximal ($M(c_j)$ defined by eq.(4)). This is repeated until the medoids converge.

$$M(c_j) = \sum_{(p_i \in c_j) \wedge (p_i \neq p_{\text{med}}(c_j))} J(p_{\text{med}}(c_j), p_i). \quad (4)$$

Other simple algorithms such as hierarchical clustering, or advanced algorithms for unsupervised learning, such as self-organizing mapping, can also be employed.

Then, we evaluate the likeliness of the records as a candidate to include unobserved persons with a ranking function $I(b_i)$. The ranking function calculates the degree of strength at which the record attracts persons belonging to multiple clusters, which originates in an unobserved person hidden in the record. The unobserved person is assumed to be a catalyst to foster the inter-cluster relationship. We present a few ranking functions. The most simple ranking function $I_{\text{av}}(b_i)$ is defined by eq.(5). It is the degree of contribution of a person $p_k$ (belonging to the cluster $c_j$) to the record $b_i$, averaged over the clusters. The records having larger value are ranked as more likely. The algorithm retrieves the records in the order of likeliness. The number of retrieved records $m^{\text{ret}}$ can be set arbitrarily (from 1 to |b|).

$$I_{\text{av}}(b_i) = \frac{1}{|c|} \sum_{0 \leq j \leq |c|-1} \max_{p_k \in c_j} \frac{B(p_k \in b_i)}{\sum_{0 \leq l \leq |b|-1} B(p_k \in b_l)}. \quad (5)$$

Eq.(5) can be converted to a simpler form in eq.(6).

$$I_{\text{av}}(b_i) = \frac{1}{|c|} \sum_{0 \leq j \leq |c|-1} \min_{(p_k \in c_j) \wedge (p_k \in b_i)} F(p_k). \quad (6)$$

A gateway person $p_{\text{gtw}}(b_i, c_j)$ in the cluster $c_j$ for the record $b_i$ is calculated by eq.(7). It is the person who maximizes the term to be averaged in eq.(5).

$$p_{\text{gtw}}(b_i, c_j) = \arg\max_k \; p_k \in c_j \; \frac{B(p_k \in b_i)}{\sum_{0 \leq l \leq |b|-1} B(p_k \in b_l)}. \quad (7)$$

Standard deviation is an alternative to calculate the likeliness. $I_{\text{sd}}(b_i)$ defined by Eq.(8) is employed instead of eq.(5) or eq.(6). The records having smaller value are ranked as more likely.

$$I_{\text{sd}}(b_i)^2 = \frac{1}{|c|} \sum_{0 \leq j \leq |c|-1} (\max_{p_k \in c_j} \frac{B(p_k \in b_i)}{\sum_{0 \leq l \leq |b|-1} B(p_k \in b_l)} - I_{\text{av}}(b_i))^2 \quad (8)$$

The average of two of the largest values Itp($b_i$) is an alternative, instead of the average over the all clusters in eq.(5) or eq.(6). This is defined by eq.(9). The records having larger value are ranked as more likely.

$$I_{\text{tp}}(b_i) = \frac{T(\max_{p_k \in c_j} \frac{B(p_k \in b_i)}{\sum_{0 \leq l \leq |b|-1} B(p_k \in b_l)}, 1) + T(\max_{p_k \in c_j} \frac{B(p_k \in b_i)}{\sum_{0 \leq l \leq |b|-1} B(p_k \in b_l)}, 2)}{2} \quad (9)$$

The function T($x_j$,k) in eq.(9) picks up the k-th element from $x_j$ sorted in descending order. More formally, it is defined recursively by eq.(10).

$$T(x_j, k) = \max_{(x_j \notin T(x_j, l)) \land (l < k)} x_j \quad (k = 0, 1, \ldots) \quad (10)$$

Finally, the retrieved records and gateway persons are visualised into a social network diagram. The unobserved person in the record bi is labelled as DEi, and drawn as a red node. The red node and the gateway persons $p_{\text{gtw}}(b_i, c_j)$ are connected with red links. A social network diagram like the inset (b) in figure 1 is drawn in this way.

## 3 Social network

We briefly review the social network responsible for the 9/11 attack in 2001 (Krebs, 2002). The study provides us with an insight on the covert social network foundation behind the terrorism disaster. The social network is also used in the simulation is section 4. (Krebs, 2002) and (Morselli, 2007) studied the social network consisting of the 19 hijackers boarding on the 4 crashed airplanes (AA11, AA77, AA175, and UA93) and the revealed 18 conspirators. The network is shown in figures 3 and 4. Figure 3 shows the hijackers. Figure 4 includes the conspirators.

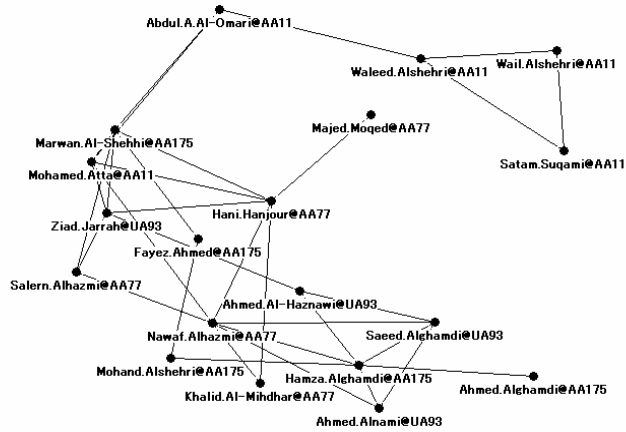

Figure 3 Social network diagram representing the observed 19 hijackers responsible for the 9/11 attack (Krebs, 2002). The flight number of the hijacked airplanes such as AA11 is shown after "@" after the hijacker names.

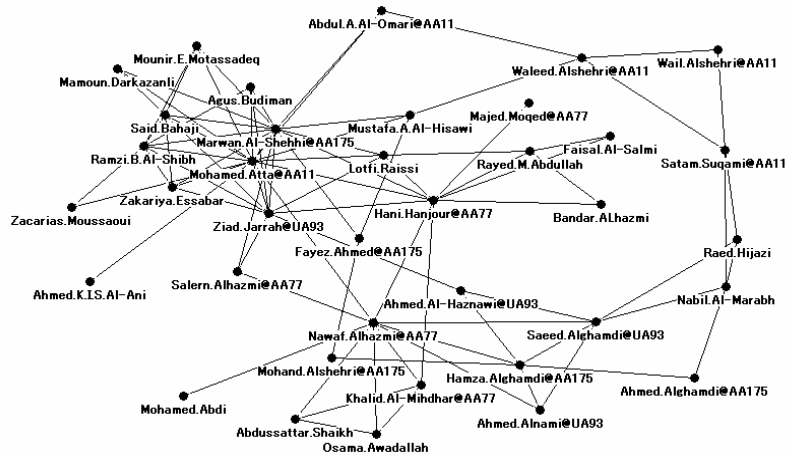

Figure 4 Social network diagram representing the observed 19 hijackers responsible for the 9/11 attack in figure 3 with the revealed 18 covert conspirators (Krebs, 2002).

The overall network topology is studied. The nodal degree averaged over the all nodes is $\mu(d) = 4.6$. Gini coefficient of the nodal degree is 0.33. The clustering coefficient averaged over the all nodes is $\mu(c) = 0.6$. It is 3.2 times larger than that in the Barabasi-albert model (Barabasi, 1999), a scale-free network, having the same Gini coefficient. Large clustering coefficient indicates that clusters exist as a core structure, but the network takes a less compact form. As qualitatively suggested by (Klerks, 2002), the terrorists possess a cluster-and-bridge structure, rather than a center-and-periphery structure. It is in agreement with the observation that the Al Qaeda network is a flexible tie-up of isolated cliques (Popp, 2006). Note that a bridge is an essential component to make clusters rendezvous to form a social network. The absence of hubs overcomes the drawbacks of a scale-free network, where the hubs result in vulnerability to attacks (Albert, 2000) and easy exposure by the efficient search over the network (Adamic, 2001).

## 4 Simulation

### 4.1 Test data

We present quantitative performance evaluation of the proposed method. The test data, as an input to our method, is communication records simulated on the 9/11 social network in section 3, and configured to include a convert conspirator as a latent structure for simulation purpose. The records are generated in the 2 steps below. In the second step, a latent structure is configured by deleting a conspirator from the records (Maeno, 2006). Note that the latent structure does not change the communication pattern in the social network, but changes observable communication.

The first step is to collect the simulated communication into records. Communication is assumed to be information dissemination over links from an initiator. It is like a conversation taking place under the subject the initiator concerns. Communication transmits on a link at a probability of t. It represents communication strength. Communication reaches $t \times \mu(d)$ persons by a hop on the average. The maximal transmission distance is limited to 2-hop long because 3-hop long communication covers most persons due to the small network size. An initiator is selected uniformly. Persons, whom communication reaches, are grouped into a record. Hijackers and conspirators are not distinguished here. The average number of persons included in a basket is $|b_i|$=6.5, 10.1, 13.7, and 17.1 at t=0.4, 0.6, 0.8, and 1.0. The number of baskets used in the evaluation is $|b|$=370. The following is example records initiated by Abdul A. Al-Omari, Mustafa A. Al-Hisawi, Waleed Alshehri, and Fayez Ahmed.

- $b_0$={Abdul A. Al-Omari, Marwan Al-Shehhi, Mohamed Atta, Waleed Alshehri}.
- $b_1$={Mustafa A. Al-Hisawi, Marwan Al-Shehhi, Mohamed Atta, Fayez Ahmed, Waleed Alshehri}.
- $b_2$={Waleed Alshehri, Abdul A. Al-Omari, Mustafa A. Al-Hisawi, Wail Alshehri, Satam Suqami}.
- $b_3$={Fayez Ahmed, Mohand Alshehri, Hamza Alghamdi}.

The second step is to configure a covert conspirator as a latent structure. A latent structure is configured to the records by deleting the conspirator (target to be inferred in the simulation) from the data. As a result, the deleted conspirator and the related links become invisible. The records, where the covert conspirator is hidden behind, are the input to the algorithm. The following is example records where Mustafa A. Al-Hisawi is configured to be a covert conspirator. The algorithm is expected to retrieve b2' and b3', which are different from b2 and b3. Such clues are used to start investigation on Waleed Alshehri who is included in both baskets.

- $b_0$'={Abdul A. Al-Omari, Marwan Al-Shehhi, Mohamed Atta, Waleed Alshehri} =$b_0$.
- $b_1$'={Marwan Al-Shehhi, Mohamed Atta, Fayez Ahmed, Waleed Alshehri}.
- $b_2$'={Waleed Alshehri, Abdul A. Al-Omari, Wail Alshehri, Satam Suqami}.
- $b_3$'={Fayez Ahmed, Mohand Alshehri, Hamza Alghamdi}=$b_3$.

*4.2 Performance evaluation*

In information retrieval, precision and recall are used as evaluation criteria. Precision p is the fraction of relevant data among the all data returned by search. The relevant data here is the records where the covert conspirator has been deleted in the second step. Recall r is the fraction of the all relevant data that is returned by the search among the all relevant data. They are defined by eq(11). and eq.(12).

$$p = \frac{\sum_{0 \leq i \leq m^{\text{ret}}-1} B(b_i' \neq b_i)}{m^{\text{ret}}} \quad . (11)$$

$$r = \frac{\sum_{0 \leq i \leq m^{\text{ret}}-1} B(b_i' \neq b_i)}{\sum_{0 \leq i \leq |b|-1} B(b_i' \neq b_i)} \quad . (12)$$

Besides, F value is useful as a geometric mean of precision and recall. It is defined by eq.(13).

$$F = \frac{1}{\frac{1}{2}(\frac{1}{p}+\frac{1}{r})} = \frac{2pr}{p+r} \quad . (13)$$

F value gain $g_F$ is defined by eq.(14). It is the ratio of the F value of the algorithm to the F value of the random retrieval.

$$g_F = \frac{F}{F_{\text{rd}}} \quad . (14)$$

Performance of the algorithm is evaluated with the test data under several conditions. Figure 5 shows precision and recall to retrieve the records where a covert conspirator, Mustafa A. Al-Hisawi, has been hidden. Mustafa A. Al-Hisawi was a big financial sponsor to the hijackers, as mentioned in section 1.The number of clusters is |c|=4. The probability of communication transmission is t=0.8. The horizontal axis is the ratio of the number of retrieved basket data to the number of the whole basket data ($m^{\text{ret}}/|b|$). The records retrieved as top 10% ranking are correct. The algorithm outputs correct information. The ranking function Isd($b_i$) seems to show a little better performance than Iav($b_i$). Isd($b_i$) is employed in the following study. Precision is 100% when the top 10% of the baskets are retrieved. The algorithm works fine. Precision is 0.45 when the all baskets are retrieved. The problem here includes many correct answers. It is not so difficult because the network is small. (Maeno, 2006) studies the performance for a network consisting of 400 nodes

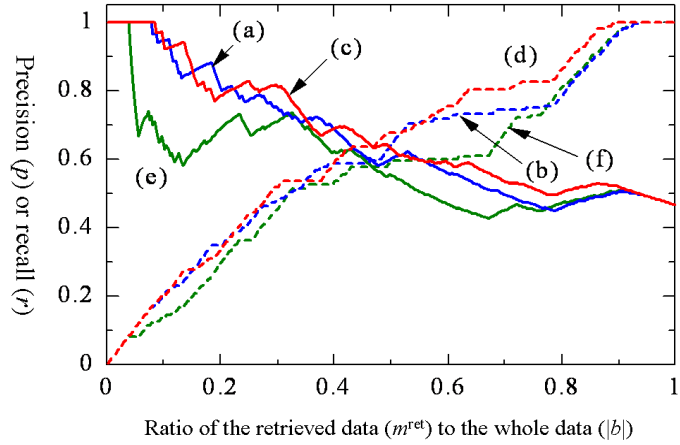

Figure 5 Precision p and recall r to retrieve the records where a covert conspirator, Mustafa A. Al-Hisawi, has been hidden: (a) p using $Iav(b_i)$, (b) r using $Iav(b_i)$, (c) p using $Isd(b_i)$, (d) r using $Isd(b_i)$, (e) p using $Itp(b_i)$, and (f) r using $Itp(b_i)$. The number of clusters is $|c|=4$. The probability of communication transmission is $t=0.8$. The horizontal axis is the ratio of the number of retrieved basket data to the number of the whole basket data (mret/|b|).

Figure 6 shows precision and recall at $|c|=2, 4, 8$, and $t=0.8$. The value of $|c|$ depends on the prior knowledge of the social network structure. The case where $|c|=4$ is a reasonable choice, based on the knowledge that 4 airplanes were hijacked. It actually shows the best performance. With the wrong prior knowledge, $|c|=2$, the performance degrades. Performance degradation at $|c|=8$ is small because the practical number of groups including conspirators may be close to, but a little larger than 4.

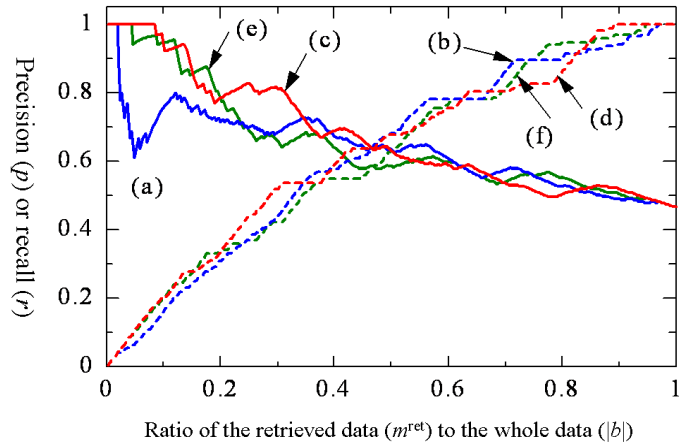

Figure 6 Precision p and recall r to retrieve the records where a covert conspirator, Mustafa A. Al-Hisawi, has been hidden: (a) p at $|c|=2$, (b) r at $|c|=2$, (c) p at $|c|=4$, (d) r at $|c|=4$, (e) p at $|c|=8$, and (f) r at $|c|=8$. The simulation condition is that $t=0.8$, and $Isd(b_i)$ is used.

Figure 7 shows F value gain at |c|=4, and t=1.0, 0.8, 0.6, 0.4. At t=1.0, 0.8, the performance is stable (the curve is smooth). At t=1.0, the gain is small because the increasing input information and longer reach communication make the problem easy. At t=0.6, the performance begins to be unstable (the curve begins to fluctuate). At t=0.4, the algorithm fails to work because the input information is too poor to extract inter-cluster relationship.

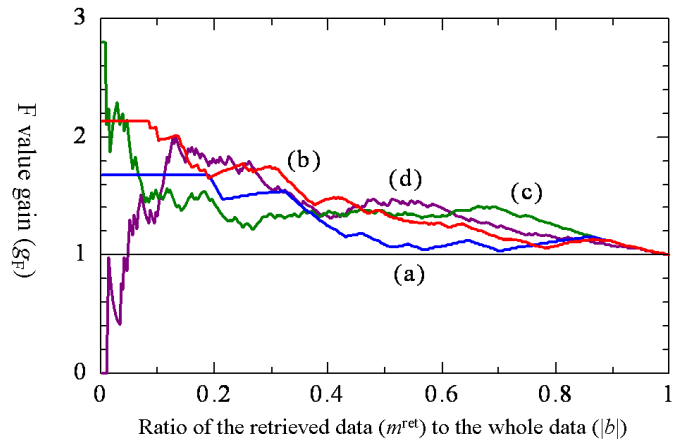

Figure 7 F value gain to retrieve the records where a covert conspirator, Mustafa A. Al-Hisawi, has been hidden: (a) t=1.0, (b) t=0.8, (c) t=0.6, and (d) t=0.4. The simulation condition is that |c|=4, and $Isd(b_i)$ is used.

Figure 8 shows F value gain for a variety of covert conspirators. The algorithm works for Lotfi Raissi, or Rayed M. Abdullah. Lotfi Raissi was under suspicion of training the pilots who hijacked the AA77 and flew it into the Pentagon. Rayed M. Abdullah trained with Hani Hanjour who hijacked the AA77. Their position in the social network is similar to Mustafa A. Al-Hisawi. For Ramzi B. Al-Shibh, or Said Bahaji, the algorithm also works, although a little degradation is observed. For Osama Awadallah, or Raed Hijazi, the performance becomes less stable and worse. Many times, Osama Awadallah met Nawaf Al-Hazmi who hijacked the AA77. Raed Hijazi was said to have connection to Osama bin Laden, and to prepare the explosives for the Millennium plot in Jordan in 2000. The degradation may arise because Osama Awadallah and Raed Hijazi are at the border of the network. Their absence in the records does not affect the overall clustering structure, and is not easy to discover. The algorithm suffers from limitation for such covert conspirators.

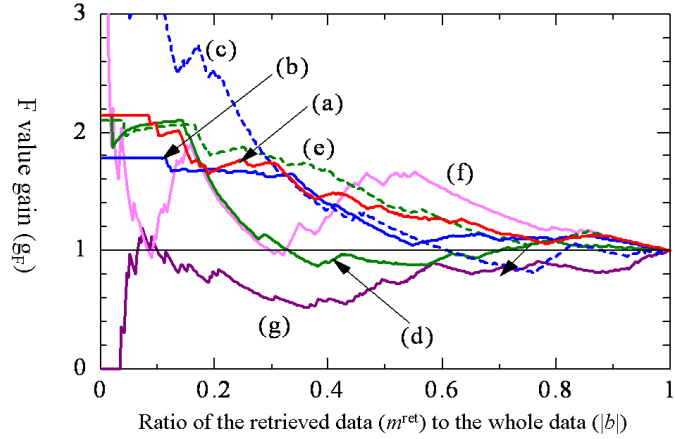

Figure 8 F value gain to retrieve the records where a covert conspirator has been hidden. The covert conspirator is (a) Mustafa A. Al-Hisawi, (b) Lotfi Raissi, (c) Rayed M. Abdullah, (d) Ramzi B. Al-Shibh, (e) Said Bahaji, (f) Osama Awadallah, and (g) Raed Hijazi. The simulation condition is that $|c|=4$, $t=0.8$, and $Isd(b_i)$ is used.

Figure 9 shows F value gain to retrieve the records where a covert conspirator, Raed Hijazi, has been hidden. $Iav(b_i)$ and $Itp(b_i)$ are employed again as in Figure 5. $Itp(b_i)$ shows better performance although it is still a little unstable and may not be sufficient for a practical use. The performance may be improved by focusing on the relationship between 2 clusters, rather than between the all clusters.

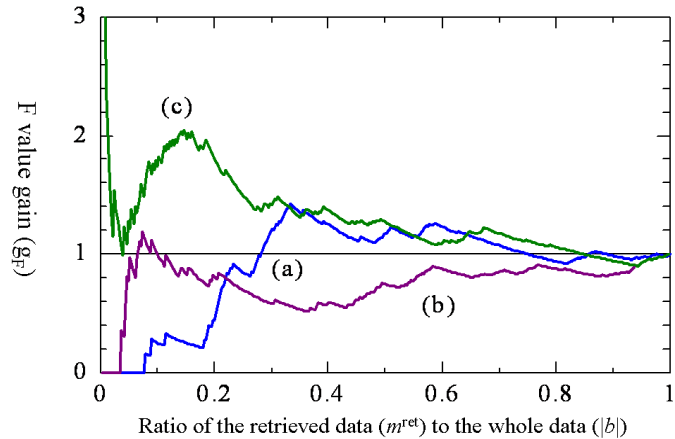

Figure 9 F value gain to retrieve the records where a covert conspirator, Raed.Hijazi, has been hidden: using (a) $Iav(b_i)$, (b) $Isd(b_i)$, and (c) $Itp(b_i)$. The simulation condition is the same as in figure 5 ($|c|=4$ and $t=0.8$).

*4.3 Social network visualization*

A social network diagram is drawn from the observed records according to the process in figure 2. The unobserved person in a suspicious record is drawn as a red node. The red node and the gateway persons $p_{\text{gtw}}(b_i, c_j)$ are connected with red links.

Figure 10 shows the social network diagram. The condition is the same as in figure 5, where a covert conspirator, Mustafa A. Al-Hisawi, has been hidden and the target to discover. The 4 terrorist groups are inter-connected with 10 of the highly ranked red nodes, DEi, corresponding to Mustafa A. Al-Hisawi hidden in the suspicious records. The bottom left cluster, including Nawaf Alhazmi, Mohamed Atta, and Hani Hanjour, is isolated and not connected to the red nodes. Terrorists who appear more frequently are less emphasized because of the denominator of eq.(5) or eq.(6). This is not a problem, but good news. We are inclined to overlook uncommon and unexpected clues by paying too much attention to something frequent and conspicuous. On the other hand, focusing on something infrequent is a double-edged sword. We may confuse the clues observed infrequently with random noise. Majed Moqed, Mohamed Abdi, and Ahmed K. I. S. Al-Ani are probably noise. They are distant from Mustafa A. Al-Hisawi in figure 4.

It is, however, remarkable that Waleed Alshehri and Mohand Alshehri are retrieved as neighbor persons of the red nodes indicating the existence of Mustafa A. Al-Hisawi. They are close to him. Waleed Alshehri, one of muscle hijackers, helped Mohammed Atta hijack the AA11 and fly it into the North Tower of the World Trade Center. Mohand Alshehri hijacked the AA175 and flew it into the South Tower of the World Trade Center. Waleed Alshehri is connected with 6 links. He is the keystone person for the investigators to gather information on relatives, friends, and associates to approach to Mustafa A. Al-Hisawi.

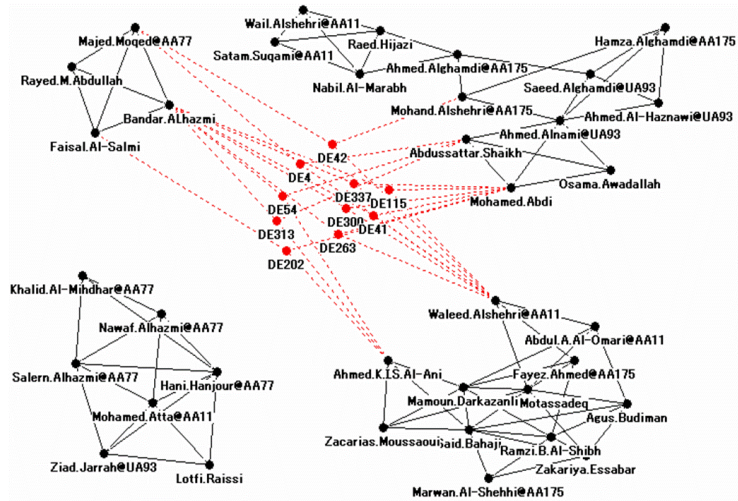

Figure 10 Four clusters and ten of the highly ranked red nodes corresponding to Mustafa A. Al-Hisawi hidden in the suspicious records. Waleed Alshehri and Mohand Alshehri are retrieved as neighbor persons of the red nodes.

## 5 Related works

Existing terrorist or criminal social networks are studied empirically. (Batallas, 2006) applied centrality (Freeman, 1979) and brokerage (Cusumano, 2000) to analyze an aircraft engine development project, and suggested relevance of an information leader team, which could be either a bottleneck or an innovation diffuser. (Keila, 2006) applied factor analysis to study email exchange in Enron, which ended in bankruptcy due to the institutionalized accounting fraud. (Klerks, 2002) points out that criminal organizations tend to be strings of inter-linked small groups that lack a central leader, but to coordinate their activities along logistic trails and through bonds of friends, and that hypothesis can be built by paying attention to remarkable white spots and hard-to-fill positions in a network. (Krebs, 2002) investigates the 9/11 terrorist network, and reveals that the relevance of conspirators who reduce the distance between hijackers and enhance communication efficiently. (Morselli, 2007) investigates Kreb's network from the viewpoint of efficiency and security trade-off, and suggests that more security-oriented structure arises from longer time-to-task of the terrorists' objectives, and that conspirators improve communication efficiency, preserving hijackers' small visibility and exposure.

Complex network, graph theory, and learning help us get an insight on the dynamics of a social network, in addition to summarizing and visualizing a network (Shen, 2007), and analyzing a cognitive network (Krackhardt, 1987). Scale-free networks (Barabasi, 1999) and small worlds (Watts, 1998) present us much insight on the structure and evolution of a social network: scientists' collaboration, actors in movies etc. A power law in the nodal degree distribution governs the scale-free network. (Fenner, 2007) proposes an exponential cutoff mechanism to modify the power law. Error attack tolerance (Albert, 2000) and search efficiency (Adamic, 2001) are of particular interest for practical applications.

Link discovery is applied to predict collaboration between scientists from the published co-authorship (Liben-Nowell, 2004). (Adamic, 2003) proposes a technique to infer friends and neighbors from the information available on the web. (Singh, 2004) applied a hidden Markov model and a Bayesian network to predict the behavior of terrorists. Learning of a Bayesian network is extended to study the probabilistic nature of latent variables. (Silva, 2006) studied learning of a structure of a linear latent variable graph. (Friedman, 1998) studied learning of a structure of a dynamic probabilistic network. The principled analytic approach often suffers from complexity problem. The complexity includes bi-directional and cyclic influence among the many observed and latent nodes (beyond a triad: 1 latent node influencing 2 observed nodes).

## 6 Concluding remark

In this paper, we demonstrate the proposed method to analyze the covert social network foundation hidden behind the terrorism disaster. The method integrates the expert investigator's prior understanding, insight on the terrorists' social network nature derived from the complex graph theory, and computational data processing. It is effective to discover a node, which functions relevantly in a social network, but escaped from monitoring on the presence and mutual relationship of nodes. Precision, recall, and F value characteristics of the algorithm are evaluated in the simulation experiment using the social network responsible for the 9/11 attack in 2001.

There are still remaining issues. How high is the quality of the hypothesis invented from the social network diagram indicating unobserved persons? We need to test the quality of the hypothesis invented by subject investigators in more realistic cases. How wide is the applicability of the algorithm in terms of social network topology, communication pattern, and their dynamical change? We need to investigate on the performance of the algorithm under more variety of environments, and to optimize the ranking function. We believe that the proposed method will contribute to understand the latent threats in social phenomena and human activities, as well as to analyze the covert social network foundation hidden behind the terrorism disaster, along with the future study for the remaining issues.